\documentclass[letterpaper, 10 pt, conference]{ieeeconf}  % Comment this line out if you need a4paper

\IEEEoverridecommandlockouts                              % This command is only needed if 
\usepackage{amsmath} 
\usepackage{amssymb}
\usepackage[mathscr]{euscript}
\usepackage{graphicx}
\usepackage{xcolor}
\usepackage{algorithm}
\usepackage[noend]{algpseudocode}
\usepackage{tabularx}
\usepackage{multirow}
\usepackage{hhline}
\usepackage{hyperref}
\usepackage[T1]{fontenc}
\usepackage{lipsum}

\usepackage[capitalise,nameinlink]{cleveref}
  \crefname{section}{Sect.}{Sect.}
  \Crefname{section}{Section}{Sections}
  \crefname{figure}{Fig.}{Fig.}
  \Crefname{figure}{Figure}{Figures}
  \crefname{table}{Tabl.}{Tabl.}
  \Crefname{table}{Table}{Tables}

\usepackage{tikz}

\newcommand\copyrighttext{%
  \footnotesize \textcopyright \the\year{} IEEE. Personal use of this material is permitted. Permission from IEEE must be obtained for all other uses, including reprinting/republishing this material for advertising or promotional purposes, collecting new collected works for resale or redistribution to servers or lists, or reuse of any copyrighted component of this work in other works.}

\newcommand\copyrightnotice{%
\begin{tikzpicture}[remember picture,overlay]
\node[anchor=south,yshift=10pt] at (current page.south) {\fbox{\parbox{\dimexpr0.75\textwidth-\fboxsep-\fboxrule\relax}{\copyrighttext}}};
\end{tikzpicture}%
}

\title{\LARGE \bf
SliceIt! - A Dual Simulator Framework for Learning Robot Food Slicing
}

\author{Cristian C. Beltran-Hernandez$^{1,*}$, Nicolas Erbetti$^{1,*}$, Masashi Hamaya$^{1}$% <-this % stops a space
\thanks{$^{*}$ Equal contribution.}
\thanks{This work was supported by KAKENHI Grant Number 21H04910.}
\thanks{
$^{1}$OMRON   SINIC   X   Corporation, Tokyo, Japan. Corresponding author
        {\tt\small cristian.beltran@sinicx.com}}
}

\begin{document}

\maketitle
\thispagestyle{empty}
\pagestyle{empty}
\copyrightnotice

\begin{abstract}
Cooking robots can enhance the home experience by reducing the burden of daily chores. However, these robots must perform their tasks dexterously and safely in shared human environments, especially when handling dangerous tools such as kitchen knives.
This study focuses on enabling a robot to autonomously and safely learn food-cutting tasks.
More specifically, our goal is to enable a collaborative robot or industrial robot arm to perform food-slicing tasks by adapting to varying material properties using compliance control.
Our approach involves using Reinforcement Learning (RL) to train a robot to compliantly manipulate a knife, by reducing the contact forces exerted by the food items and by the cutting board. However, training the robot in the real world can be inefficient, and dangerous, and result in a lot of food waste. Therefore, we proposed SliceIt!, a framework for safely and efficiently learning robot food-slicing tasks in simulation. Following a real2sim2real approach, our framework consists of collecting a few real food slicing data, calibrating our dual simulation environment (a high-fidelity cutting simulator and a robotic simulator), learning compliant control policies on the calibrated simulation environment, and finally, deploying the policies on the real robot.
% Mention somehow the performance of our system
\end{abstract}

%%%%%%%%%%%%%%%%%%%%%%%%%%%%%%%%%%%%%%%%%%%%%%%%%%%%%%%%%%%%%%%%%%%%%%%%%%%%%%%%
\section{INTRODUCTION}

% Focus on cooking
Cooking robots, which can safely work alongside humans, hold the potential to enhance home environments and ease daily chores. Tasks such as food slicing require the robot to skillfully and safely manipulate a knife. Our research focuses on enabling a robot to learn food-cutting tasks.

% RL and simulation
Cutting skills such as chopping and slicing, require manipulating the knife and carefully responding to the reaction forces that are exerted by the material and by the cutting board~\cite{mu2019robotic}. In particular, it is important for the robot to be able to adapt to the widely varying physical properties of each food product~\cite{lenz2015deepmpc}. 
Learning-based methods are promising approaches to autonomously learning complex robotic behaviors, such as the one required for food slicing. However, such methods often need a large number of interactions to learn useful control policies, which for the food-slicing tasks could result in a lot of food waste. Furthermore, it could be dangerous to train the robot directly in the real world when the robot is handling hazardous tools such as a kitchen knife, as part of its learning paradigm involves the random exploration of actions. Thus, learning in a simulation environment stands out as a viable solution, where exploratory actions can be conducted safely.

% Reality gap
Learning in simulation has its own challenges, namely the reality gap~\cite{jakobi1995noise}. The difference between the simulation environment and the real world can render the learned control policy unusable in the real world. 
Recently, advanced simulators like DiSECt~\cite{heiden2023disect} have been introduced, offering a highly realistic representation of soft material cutting. These simulators can potentially bridge the reality gap. However, to facilitate more accessible evaluation in real-world scenarios, creating an interface between the specialized cutting simulation and real robots is essential. 
The existing simulators focus on the physical interactions between the knife and the object but do not consider whether the robot can feasibly realize the knife motion. 
%We propose a dual simulation environment that combines the cutting simulator with a robotics simulator. The advantage of using a robotics simulator is that it incurs a much lower computational cost.

\begin{figure}
    \centering
    \includegraphics[width=0.4\textwidth]{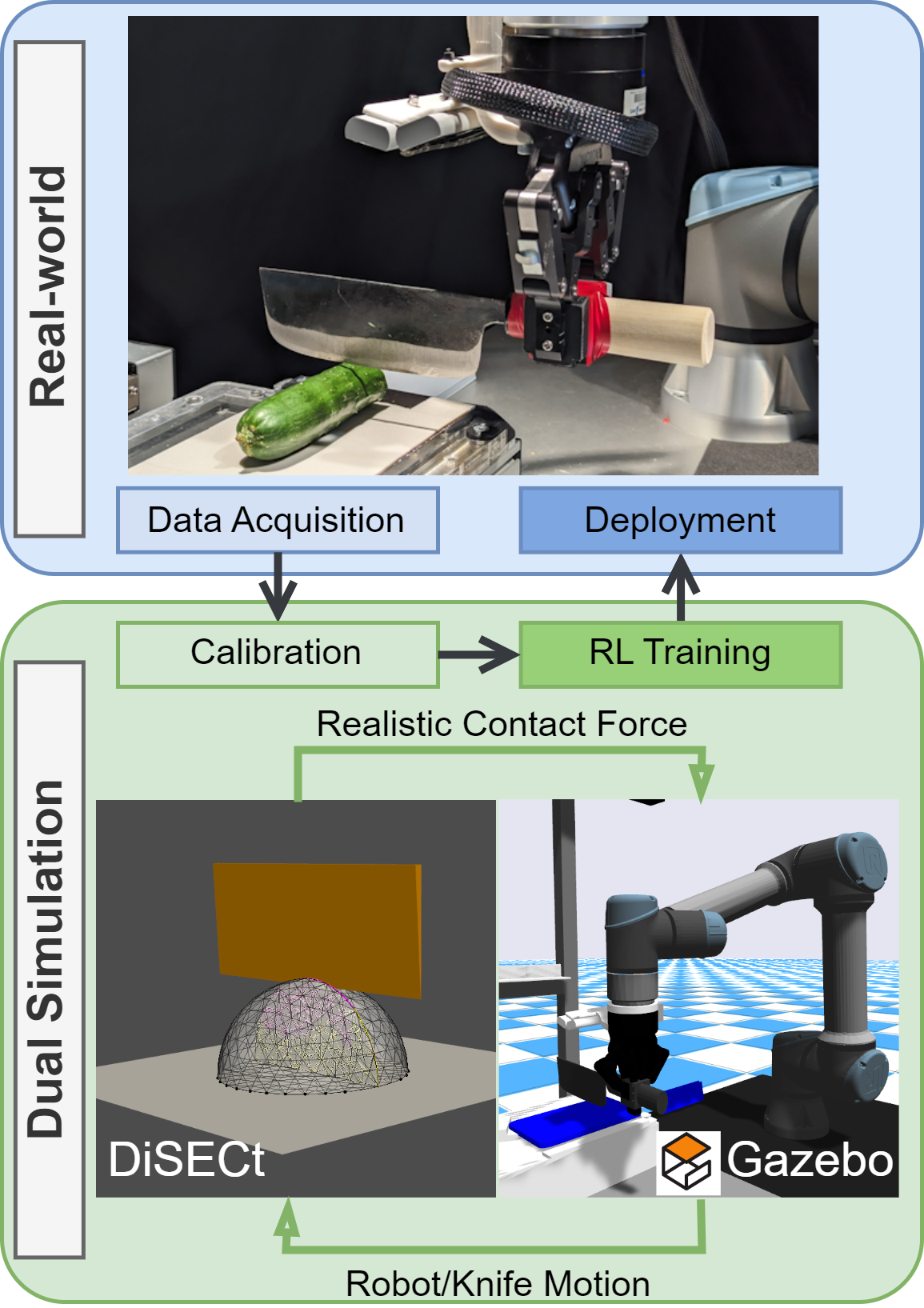}
    \caption{Overview of proposed learning-based robot cutting framework, comprising four key stages: 1) Data collection on the real robot. 2) Calibration of the cutting simulator, DiSECt. 3) Learning a control policy within a dual simulation environment using Gazebo and DiSECt. 4) Deployment to the real robot.}
    \label{fig:system-overview}
\end{figure}
% Contributions overview
To address this challenge, we propose a dual simulation environment combining the cutting simulator (CutSim) with a robotics simulator (RoboSim). The CutSim provides more realistic interaction forces to the RoboSim. Meanwhile, the RoboSim provides the knife motion generated by the simulated robot to the CutSim. Additionally, the RoboSim is more practical to use since the motions generated can be more easily integrated into the real robot~\cite{gazebo2004}.
In this work, we present SliceIt!, a framework for safely learning robot cutting, that features a dual simulation environment. Our system follows a real-to-simulation-to-real (real2sim2real)~\cite{chang2020sim2real2sim} formulation and consists of (1) data collection from slicing real food items. (2) Calibration of the CutSim to simulate with high fidelity the collected slicing data. (3) Learning a control policy using our calibrated dual simulation environment. (4) Deployment of the policy on the real robotic system. An overview of our proposed method is depicted in \Cref{fig:system-overview}.

% Contributions
The main contribution of this work is SliceIt! - A framework for safely and autonomously learning robot food-slicing tasks in simulation. To develop this system, we introduce the following technical contributions:
\begin{itemize}
    \item A dual simulation environment, that combines a high-fidelity cutting simulator and a robotic simulator, to enable better sim2real transfer of learning-based control policies. SliceIt! produced slicing motions with smaller contact forces than the baseline, where only RoboSim is used, thanks to the realistic simulation obtained from CutSim.
    \item A deep RL-based compliance control method for robotic cutting tasks, in particular for rigid robot manipulators. Our method adapts to unseen objects.
\end{itemize}
Our method was evaluated on a real robotic system and against a baseline that considers a simpler cutting simulation. Additionally, our project is open source\footnote{Available at \url{https://github.com/omron-sinicx/sliceit}} for the benefit of the research community.
% and will be available at \url{github.com/omron-sinicx/sliceit}.

This paper is organized as follows; related work is discussed in \Cref{sec:related-work}, a detailed description of SliceIt! is provided in \Cref{sec:methodology}, experiments and conclusion are provided in \Cref{sec:experiments} and \ref{sec:conclusion}.

\section{RELATED WORK} \label{sec:related-work}

\subsection{Slicing Robots} % research works that are the most similar to ours

% Model-based control
Prior works have formulated model-based control methods for robotic cutting based on analysis of stress and fracture forces of bio-materials, and blade sharpness and slicing angle~\cite{pressingslicing2006, razorcut2011}, as well as, considering cutting force models of the knife "pressing and slicing"~\cite{mu2019robotic}. 
Additionally, multimodal haptic sensory data has been used to enhance the cutting robotic system; in ~\cite{imagemoments2014} a force/visual control approach is proposed while the use of tactile sensors was used to avoid slippage of the knife during cutting in ~\cite{yamaguchi2016combining}.
To handle a wider diversity of food products, machine learning-based approaches have been proposed based on deep model predictive control~\cite{lenz2015deepmpc, mitsioni2019data} and learning from demonstrations~\cite{yang2018dmps, anarossi2023deep}, reinforcement learning~\cite{padalkar2020learning}. However, these methods require several or many real-world interactions to collect data, which for food-slicing tasks, it may result in unnecessary food waste.
Recent notable studies used sim-to-real transfer learning approaches. In RoboNinja~\cite{xu2023roboninja}, a multi-material cutting simulator is proposed to collect demonstrations to train a learning-based cutting method. DiSECt aimed at leveraging its differentiable simulation to optimize the motion of the knife. In their research, the optimization is focused on the parameters of a sinusoidal trajectory, the amplitude, downward velocity, and frequency of the slicing motion.
% What we offer
In this study, we proposed a framework to handle more complex slicing motions by learning compliant control cutting motions using deep reinforcement learning in a real2sim2real formulation. Our method requires as few as one slice motion of each food item in the real world to obtain the force profile to optimize Cutsim. Besides, our trained policy adapts to unseen objects. 

% Roboninja used reinforcement learning to teach a robot arm to extract a rigid core out of a soft material, such as avocado. DiSECt aimed at leveraging its differentiable simulation to optimize the motion of its robot. Their research optimized the amplitude, downward velocity, and frequency of the slicing motion~\cite{heiden2023disect}. 

% However, these studies mainly focus on a predetermined sinusoidal motion. With a force-compliant controller trained within our RL environment, we propose to explore the full possibilities of the 3 degrees of freedom of the robot in the cutting plane. Besides, we developed and released an interface to connect DiSECt to real robots.

% RobotNinja: Cutting simulation of multi-material. Knife motion was generated using a learning-based method using demonstrations from the simulator.
% Disect: Cutting simulation. Knife motion optimization focused only on sinusoidal motions.

\subsection{Open-Source Software for Robot Manipulation}
%Benchmarks are essential for quantitative comparison between alternative methods~\cite{calli2015benchmarking}.
Recent studies offer open-source software for various robot manipulation, for example, grasping~\cite{calli2015benchmarking}, learning~\cite{wolczyk2021continual,james2020rlbench, delhaisse2020pyrobolearn, zhu2020robosuite}, assembly~\cite{collins2023ramp}, cloth manipulation~\cite{clark2023household}, and human-robot interaction~\cite{mower2023ros}.
We present an open-source software for robotic cutting that has not been explored in the existing studies.
% What we offer
It provides an interface to integrate a dual simulation environment with real robots, making it practical to evaluate real-world scenarios.
Given the hazards associated with cutting tasks, our approach offers the safety of training the robot in a simulated environment, and a collision-averse learning method based on \cite{beltran2020variable}, which learns adaptive compliant control policies to minimize high contact forces.

% Given the potential hazards associated with cutting tasks, we implement a position-based compliant controller and some operations, such as AAA and BBB, based on~\cite{beltran2020variable} to to guarantee a secure cutting.
% It has a robot interface that connects a dual simulator and the robot, making the system immediate use. 

\subsection{Multiple Simulator}
Eaton et al \cite{eaton2016bridging} proposed using two simulators to compensate for the sim-to-real gap. One simulator was used first to train the policy, and then the policy was transferred to another simulator to evaluate its performance in a different environment. The successful policies in both simulators could be transferred to the real world. Unlike this study, which evaluated the robustness of the trained policy using different simulators, we propose a different use of the dual simulation to leverage the strengths of both simulators: the precise contact physics in CutSim (DiSECt) and RoboSim (Gazebo).
In our study, we used DiSECt as our chosen cutting simulator. However, it's important to note that our method is not limited to this specific simulator. Our approach can be adapted to other simulators, like RoboNinja~\cite{xu2023roboninja}, to handle different complex cutting tasks.

% Multiple physics simulators are often used to compare their ability to cope with the reality gaps~\cite{collins2019quantifying, acosta2022validating}. In this study, we compared the performances of only using the Gazebo~\cite{gazebo2004} and our proposed dual simulation environment Gazebo-Disect in robotic cutting tasks.

\section{METHODOLOGY} \label{sec:methodology}

% Re-organize the Methodology section with the following structure

%  System Overview: We propose a robot learning system for food-slicing tasks that uses a real2sim2real strategy. We collect one or a few samples of force profiles for slicing food items. These samples are used to calibrate the simulation environment. Here we proposed using the DiSECt simulator. Then, a robot control policy is learned in the calibrated simulation environment, using a twin-simulator setup. Finally, the slicing policy is evaluated on the real robot.

%  Data collection: Brief explanation of how the data was collected. System configuration, robots, sensors, constant speed, conversion to data format on DiSECt

%  Calibration: Describe Disect and the calibration process. The proposed optimization process.

%  Policy learning: Describe the Twin environment. Then describe the learning process.

\subsection{System Overview}
% Explain roughly all the components of the system, better with an image and a bit of text

We introduce a robot learning system designed for food-slicing tasks, employing a real2sim2real~\cite{chang2020sim2real2sim} approach. In \Cref{fig:system-overview}, we outline the core elements of our proposed method.

First, the "real2sim" phase involves data acquisition and calibration of our dual simulation environment. Our simulation environment consists of two concurrent simulators: a physics simulator tailored for cutting soft materials (CutSim), and a robotic simulator (RoboSim). 
We gather data by having the real-world robot slice food items. This data is then utilized to fine-tune the simulation parameters of the cutting simulator. Only a few data samples are required for the calibration.

Second, the "sim2real" phase focuses on training a control policy using Reinforcement Learning (RL) within the simulation and deploying it in the real world. The combined cutting simulator and the robotic simulator are used for this purpose.

The following sections describe in detail the components of our method, the dual simulation environment, the calibration of our simulation environment, and the learning-based compliance control policy.

% We propose a robot learning system for food-slicing tasks that uses a real2sim2real approach. \Cref{fig:system-overview} describes the main components of our proposed method. First, the \textit{real2sim} part refers to the data acquisition and calibration of our simulation environment. Our simulation environment consists of two simulators running in parallel; one is DiSECT \cite{heiden2023disect}, a physics simulator for cutting soft materials, and the other one is a robotic simulator. The force profile of slicing food items is collected using the real-world robot; just one or a few data samples are necessary. This data is then used to calibrate the simulation parameters of the DiSECt simulator. Second, the \textit{sim2real} part refers to the learning of a control policy using RL in simulation and deploying it to the real world. We propose using our dual-sim environment for this purpose.

% The following sections describe in detail the components of our system, the data collection of force profiles, the calibration of our simulation environment, and the policy learning.

\subsection{Dual Simulation Environment}
\begin{figure}
    \centering
    \includegraphics[width=0.4\textwidth]{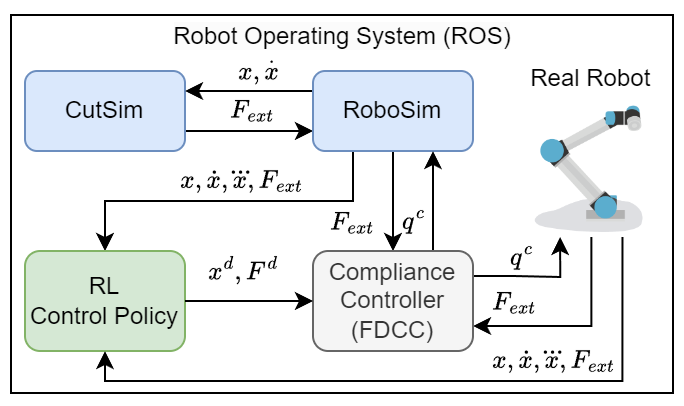}
    \caption{ROS-powered proposed system for learning robotic cutting tasks using a dual simulation environment, reinforcement learning, and compliance control.}
    \label{fig:system-diagram}
\end{figure}

% To train a robot to solve cutting tasks using simulation, we propose using a dual simulation environment consisting of a physics simulator for cutting soft materials and a robotic simulator.

% - Explain what and why is disect.  - Explain the what and why calibration process. - Explain what and why gazebo.
%
%
\subsubsection{CutSim}
    
    % What is Disect?
    In this work, the DiSECt simulator was used. DiSECt is a differentiable physics simulator for cutting soft materials~\cite{heiden2023disect}. The simulator augments the finite element method (FEM) with a continuous contact model based on signed distance fields (SDF), as well as a continuous damage model that inserts springs on opposite sides of the cutting plane and allows them to weaken until zero stiffness to model crack formation. 
    DiSECt was chosen because it allows us to realistically simulate food cutting by calibrating its simulation parameters. The differentiability of the simulator enables the calibration of the simulation parameters using gradient-based optimization methods~\cite{heiden2023disect}. 

    \textit{Calibration}: The calibration process involves simulating the robot's cutting actions, including motion and contact force, and adjusting the simulation parameters until the force profile matches the desired one. The simulator's differentiability allows us to fine-tune these parameters using gradient-based optimization methods~\cite{heiden2023disect}. However, gradient-based optimization can be computationally intensive, and inappropriate initial parameters sometimes cause learning instability. Therefore, in this study, we propose a two-step approach. Initially, a non-gradient-based optimization method is used to quickly and cost-effectively identify an initial set of simulation parameters. Subsequently, a gradient-based optimization method, specifically the Adam algorithm \cite{adam2015}, is employed to further refine these simulation parameters. To optimize the initial simulation parameters, we utilize the Tree-structured Parzen Estimator algorithm~\cite{bergstra2011algorithms} as implemented in Optuna~\cite{optuna2019}.

\subsubsection{RoboSim}
    We use the Gazebo simulator, which is an open-source 3D robotics simulator~\cite{gazebo2004}. Our choice of the Gazebo simulator was motivated by its compatibility with the Robot Operating System (ROS)~\cite{quigley2009ros}. ROS is an open-source robotics middleware that facilitates the integration of different robotic components. In this work, ROS was used to integrate both simulators, the real robots, our proposed RL control policy, and its low-level compliance controller as depicted in \Cref{fig:system-diagram}. 

\subsubsection{Simulators' Bridge}
    Our decision to employ two simulators is driven by the high computational cost of CutSim. To obtain better results from CutSim, the simulation needs to run at a much higher frequency. In practice, a time step of $dt=1.0e^-5$ gave us the best results. Meanwhile, RoboSim can run at a lower frequency, thus requiring much less computational cost. A time step of $dt=1.0e^-4$ was used for RoboSim. However, RoboSim cannot consider the interaction forces in the cutting plane, it can only consider the normal contact force against the surface of objects.

    In our proposed framework, both simulators run in parallel with an interleaving simulation of a set time duration. In practice, we use the highest control frequency of the Universal Robots UR5e, which translates to a duration of $2.0e^{-3}$~seconds. The simulators exchange information through ROS messages. RoboSim provides the position $x$ and velocity $\dot x$ of the knife with respect to the robot while CutSim provides the contact force $F_{ext}$ for the robot compliance controller, as depicted in \Cref{fig:system-diagram}. 

\subsection{Learning slicing with reinforcement learning} 
\subsubsection{Markov Decision Process}
% Explain the RL framework used (TODO: Cristian)
    In RL, agents learn a desired behavior through interaction. The slicing tasks can be formulated as an episodic Markov Decision Process (MDP) that has finite time steps with a limit of $T$ steps per episode. MDP can be denoted as a tuple $(\mathcal{}{S}, \mathcal{A}, \mathcal{P}, \mathcal{R}, \gamma)$, where $\mathcal{S}$ is the state space, $\mathcal{A}$ is the action space, $\mathcal{P}:\mathcal{S} \times \mathcal{A}$ is a transition probability function, $\mathcal{R}$ is a reward function, and $\gamma \in (0,1)$ is a discount factor.
    At each time step $t$, the agent observes the current state $\textbf{s}_t \in \mathcal{S}$ and takes an action space $\textbf{a}_t \in A$ according to a policy $\pi$. The environment changes to the state $\textbf{s}_{t+1}$ according to the transition function $p(\textbf{s}_{t+1} \,|\, \textbf{s}_t, \textbf{a}_t)$. Then, the agent receives a numerical reward $\textbf{r}_t = R(\textbf{s}_t, \textbf{a}_t, \textbf{s}_{t+1})$ for taking action $\textbf{a}_t$ at the state $\textbf{s}_t$ and landing to state $\textbf{s}_{t+1}$. The goal is to find a policy $\pi^*$ that maximizes the expected sum of discounted future rewards given by $R(t)=\sum_i^{T} \gamma \textbf{r}_t$ \cite{sutton2018reinforcement}.

    We used the RL algorithm soft actor-critic (SAC). SAC \cite{haarnoja2018soft} is an off-policy actor-critic deep RL algorithm based on maximal entropy. SAC aims to maximize the expected reward while also optimizing maximal entropy, $H$. The agent optimizes a maximal entropy objective, encouraging exploration according to a temperature parameter $\alpha$. 

    \begin{equation*}
        \pi^* = arg \max_{\pi} E_{\pi} \left[\sum_{t=0}^{T} \gamma^t \Bigl( \mathcal{R}(\textbf{s}_t, \textbf{a}_t, \textbf{s}_{t+1}) + \alpha H(\pi (\cdot|\textbf{s}_t) \Bigr) \right]
    \end{equation*}
    
    The core idea of this method is to succeed at the task while acting as randomly as possible. SAC is an off-policy algorithm that uses a replay buffer to reuse information from recent rollouts for sample-efficient training. Thus, SAC can be enhanced with the prioritized experience replay \cite{schaul2015prioritized} approach for further improvement.

\begin{figure}[t]
    \centering
    \includegraphics[width=0.45\textwidth]{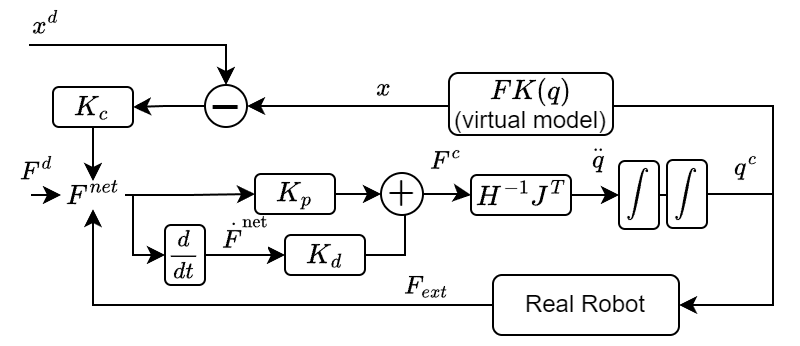}
    \caption{Forward Dynamics Compliance Controller \cite{scherzinger2017forward}}
    \label{fig:fdcc}
\end{figure}
\subsubsection{Compliance control} \label{sec:compliance-control}

For the contact-rich task of food slicing with a rigid robotic arm, the Forward Dynamics Compliance Control (FDCC) method \cite{scherzinger2017forward} was used. FDCC combines three control principles: Impedance Control, Admittance Control, and Force Control into one new strategy to realize Cartesian compliance control. As a key component in FDCC, forward dynamics simulations of a virtual model are leveraged to directly map Cartesian inputs to joint control commands, leading to good stability in singularities. Our proposed system uses the open-source implementation \footnote{FZI Cartesian Controllers: \\ github.com/fzi-forschungszentrum-informatik/cartesian\_controllers} of FDCC for ROS.

The controller is described in \Cref{fig:fdcc}, where $x$ represents the Cartesian position of the robot's end-effector and the joint positions. $F_{ext}$ is the measured contact force. $x^d$ and $F^d$ are the desired pose and wrench of the robot's end-effector, and $F^c$ is the commanded wrench. The virtual model's forward dynamics and forward kinematics are $H^{-1}J^T$ and $FK(q)$, respectively. $q$ and $\ddot q$ correspond to the position and acceleration of the robot joints. The net force $F^{net}$ encompasses the target forces, the forces measured by the sensor, and all the forces that result from virtual motion. The controller is parameterized by the stiffness $K^c$, and the PD gains $K^p$ and $K^d$. 

\subsubsection{RL agent}

\begin{figure}[t]
    \centering
    \includegraphics[width=0.48\textwidth]{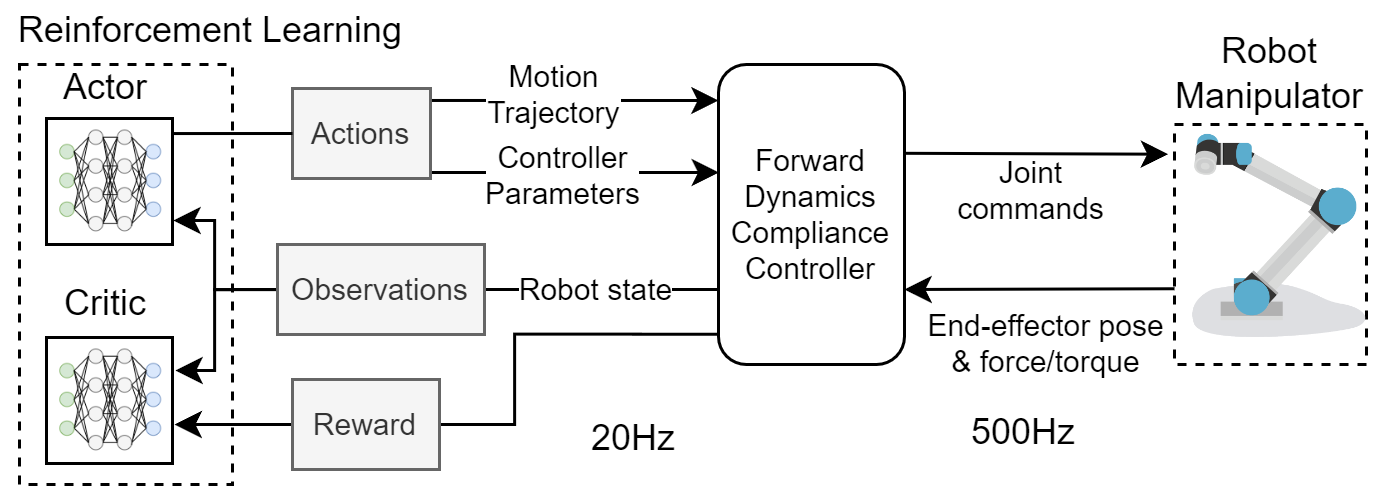}
    \caption{Framework for learning food slicing using RL and compliance control. The RL agent controls provides the reference trajectory and control parameters for the compliance controller.}
    \label{fig:compliance-control}
\end{figure}

Traditionally, fine-tuning a compliance controller for a given task is a time-consuming process that requires human expertise. To reduce these requirements, we use RL to learn the motion of the cutting action as well as the control parameters of the FDCC, inspired by previous work \cite{beltranhern2020learning}. Compared to \cite{beltranhern2020learning}, we use FDCC because it provides better stability in singularities and requires fewer parameters to learn.
As described in \Cref{fig:compliance-control}, the actions of the RL agent provide the reference trajectory $x^c$ and the control parameters $[K^c, K^p, K^d]$ to the FDCC at a low control frequency. Then the FDCC directly controls the robot with joint commands at the highest control frequency available. The feedback from the robot is the knife pose, computed using forward kinematics from the robot's joint positions, and the sensed force and momentum. The agent's observations consist of the knife's position relative to a target position, its velocity and jerkiness, as well as the previous action taken and the history of $n$ force-torque readings from the sensor. 

The SAC's actor-network architecture consists of a Temporal Convolutional Network (TCN) \cite{BaiTCN2018}, that processes a history of $n$ force/torque readings from the sensor, and a fully connected network that processes the remaining observations. Both networks output 64 features that are concatenated and processed together on an additional fully connected network, a simplified version of this structure is shown in \Cref{fig:compliance-control}. The choice of network architecture is based on results from previous work \cite{beltran2020variable}.

\subsubsection{Reward Function}
The reward function is defined based on the height of the knife with respect to the cutting board $x_{cut}$, the contact force $F_{ext}$, and the jerkiness of the knife $\dddot x$. Additionally, we defined reward values for conditions that terminate the episode, completing the task (the knife reaches the desired target pose within some threshold), colliding with the cutting board (exceeding a maximum contact force), moving outside of the defined workspace, and taking too long to complete the task.

\begin{equation}
    r = w_1\tanh(|x_{cut}|) - w_2/(1+e^{||F_{ext}||}) - w_3||\dddot x|| + \mathfrak{P}
\end{equation}
where $\mathfrak{P}$ is 
\begin{equation*}
    \mathfrak{P} = \left\{\begin{matrix}
     100,    & \textrm{Task completed}\\ 
     -100,   & \textrm{Collision} \\
     -1,     & \textrm{Otherwise}
    \end{matrix}\right.
    \label{eq:safety-reward}
\end{equation*}

\section{Experiments And Results} \label{sec:experiments}

The following experiments were designed to answer these questions:
\begin{itemize}
    \item How well does our system perform in the real world when it has only been trained in a simulated environment?
    \item When it comes to the robotic cutting task, does using a more realistic simulation environment lead to better performance compared to using a less detailed one?
\end{itemize}

For the latter, we evaluate the performance of our proposed system against a baseline. This baseline involves calibrating the simulation environment and training a policy exclusively using the Gazebo Simulator~\cite{gazebo2004}. 

\begin{figure}[t]
    \centering
    \includegraphics[width=0.4\textwidth]{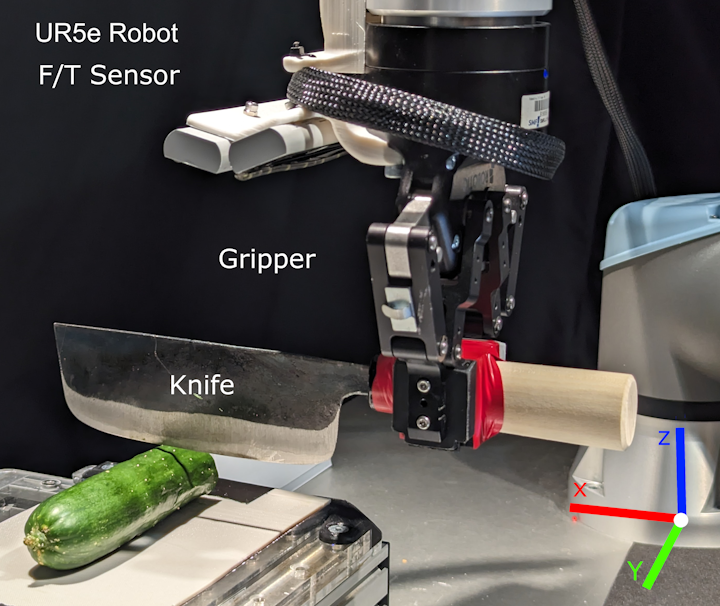}
    \caption{Experimental setup with the real robot.}
    \label{fig:real-setup}
\end{figure}
\begin{figure*}[t]
    \centering
    \includegraphics[width=0.95\textwidth]{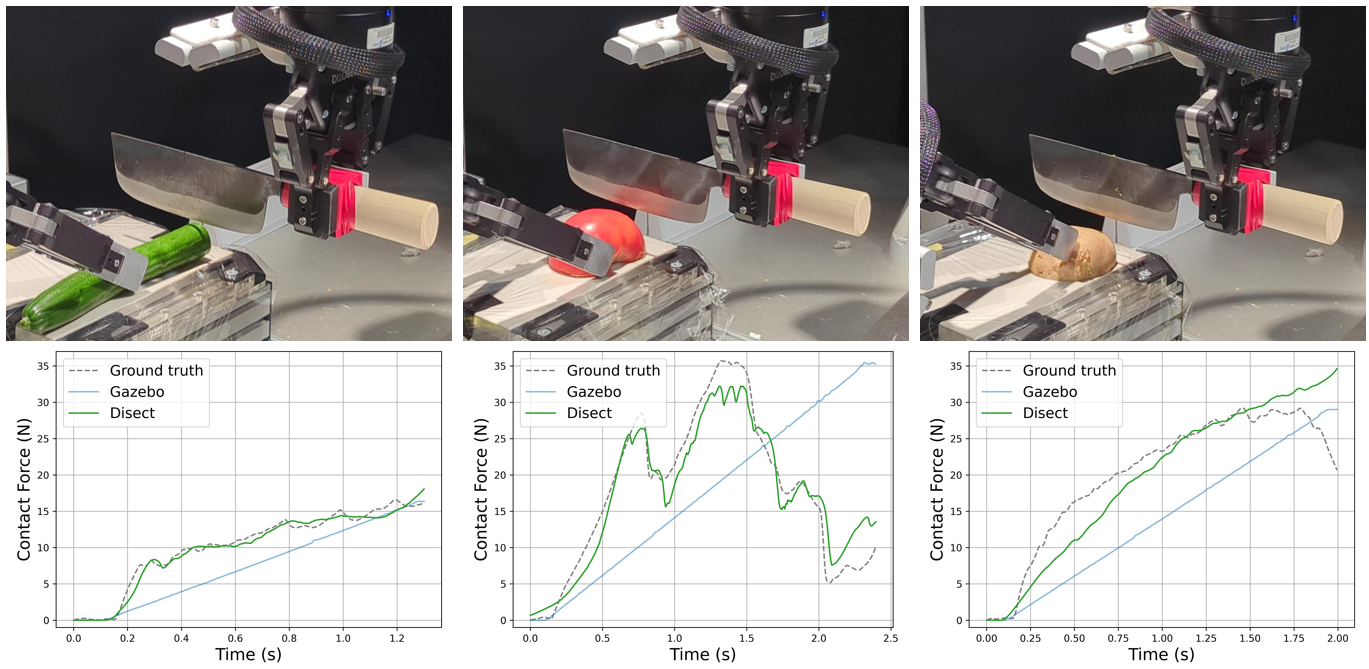}
    \caption{Force profile of the simulated cutting after the calibration process for both simulators. }
    \label{fig:force_profile}
\end{figure*}
\subsection{Experimental Setup}
\subsubsection{Robotic Platform}

Our robotic platform is based on the dual-arm system initially introduced in~\cite{von2022team}. It comprises two Universal Robot UR5e robotic arms, each equipped with a parallel gripper and a force-torque sensor positioned at the arm's end. In the context of this research, one of these robots serves a supporting role, such as holding the vegetable, while the second robot is responsible for executing the cutting task.  The experimental setup is depicted in \Cref{fig:real-setup}. Notably, the robot engaged in the task is equipped with a Robotiq 85 parallel gripper featuring a custom finger adapter that enables the attachment and detachment of a kitchen knife.

\subsubsection{Calibration}

In this study, we gathered data from three distinct food items, each characterized by different material properties, to ensure diversity within our training dataset. Specifically, we selected a tomato, a cucumber, and a potato, representing items with low, medium, and high stiffness, respectively. The collected data consists of the knife's motion and the contact force applied during the cutting action. For simplicity, we maintained a constant knife speed throughout the experiments. The simulation parameters of the Disect simulator considered during the calibration process are detailed in \Cref{tab:calibration-parameters}.

In the baseline case, denoted as \textit{Gazebo only}, the cutting action was simulated using a compression spring. In the simulator, the compression spring is defined as a prismatic joint where the force constant is determined by specific simulation parameters, namely, the Error Reduction Parameter (ERP) and Constraint Force Mixing (CFM). These parameters were calibrated so that the force required to compress the spring to its maximum matches the maximum force observed in the real-world force profile, as depicted in \Cref{fig:force_profile}.

\subsubsection{Cutting Task}
% Describe the RL task, goal, and conditions
In this study, the robot cutting task is defined as a single slice of the food item, which can then be executed multiple times as necessary. The goal of the robot is to maximize speed and minimize contact force and motion jerkiness. In particular, the task includes minimizing the force exerted on the cutting board with the knife. Four food items were used for validation: a cucumber, a tomato, a potato, and a carrot. As mentioned above, the first three were used for calibrating the simulation environments, while the carrot was used to evaluate the generalization capabilities of our method. 

After calibration of both our method and the baseline using all of the food items, an RL agent was trained in each simulation environment for 80K time steps. The training included domain randomization by loading one of the calibrated food items into the CutSim as well as injecting a small uniformly sampled noise into the simulation parameters. A similar noise was injected into the baseline. The learning converges at around 60K time steps.

\begin{table}[t]
\caption{Calibration parameters for the DiSECt simulator.}
\label{tab:calibration-parameters}
\centering
\begin{tabular}{|c|c|c|}
\hline
Parameter & Range \\
\hline
Cut Springs' stiffness & [100, 8000] \\
Cut Springs' softness & [10, 5000] \\
Contact stiffness & [200, 8000] \\
Contact damping & [0.1, 100] \\
Contact friction stiffness & [0.001, 8000] \\
Contact friction coefficient  & [0.45, 1.0] \\
\hline
\end{tabular}
\end{table}

\subsection{Real World Experiments}

The evaluation of the learned control policies involved slicing each food item multiple times. To minimize food waste, only one unit of each food item was utilized in these experiments. The specific experimental conditions are presented in \Cref{tab:experimental_conditions}.

\begin{table}[t]
\caption{Real-world experimental conditions}
\label{tab:experimental_conditions}
\centering
\begin{tabular}{|c|c|c|}
\hline
Food Item & \# of Slices & Slice size (mm) \\ \hline
Cucumber  & 15           & 5               \\ 
Tomato    & 5            & 5               \\ 
Potato    & 10           & 3               \\ 
Carrot    & 5            & 5               \\ \hline
\end{tabular}
\end{table}

\begin{figure}[t]
    \centering
    \includegraphics[width=0.45\textwidth]{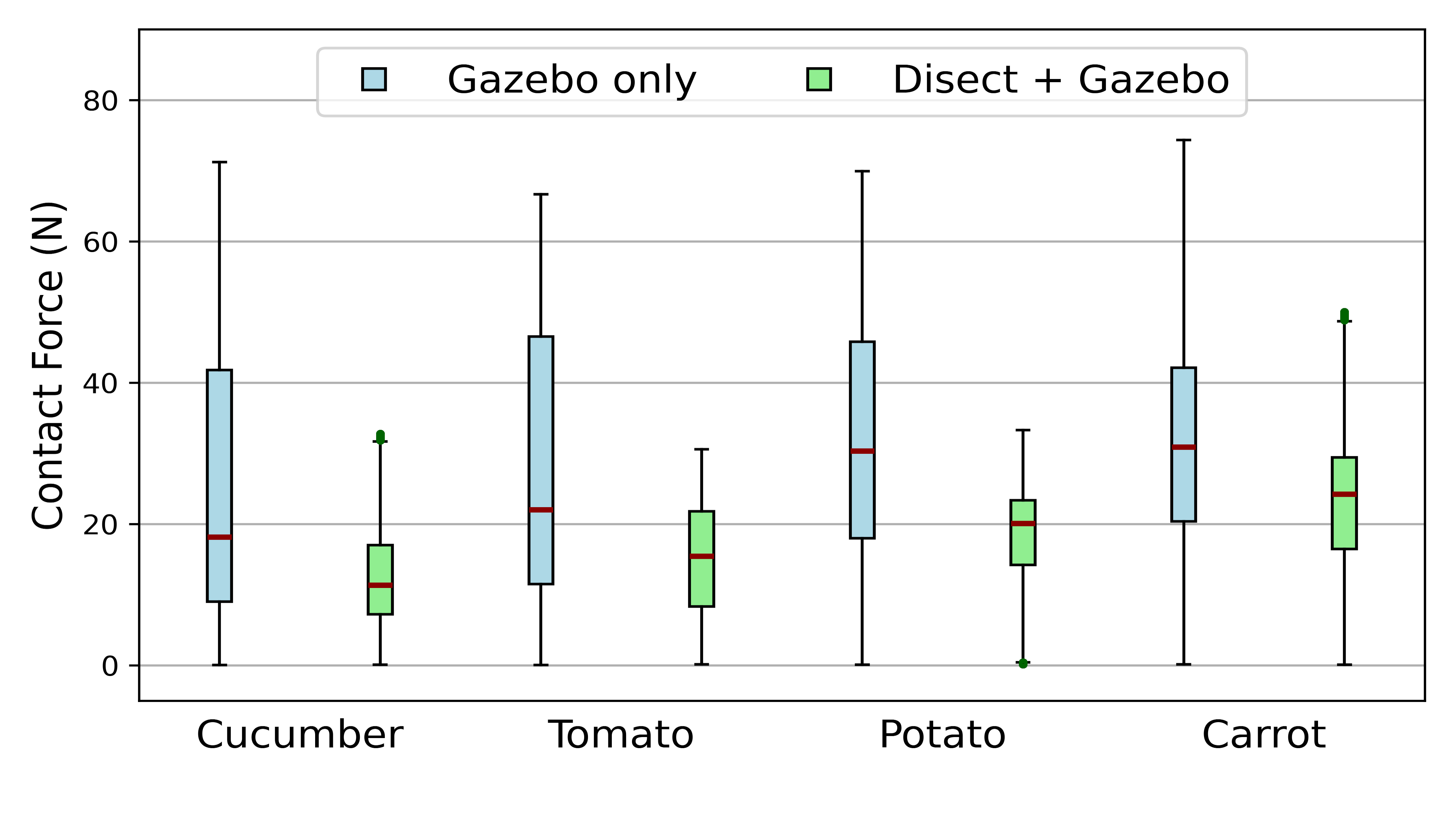}
    \caption{Evaluation on the real robot: Slicing contact force for four vegetables, cucumber, tomato, potato, and carrot.}
    \label{fig:exp-boxplot}
\end{figure}
\begin{figure}[t]
    \centering
    \includegraphics[width=0.45\textwidth]{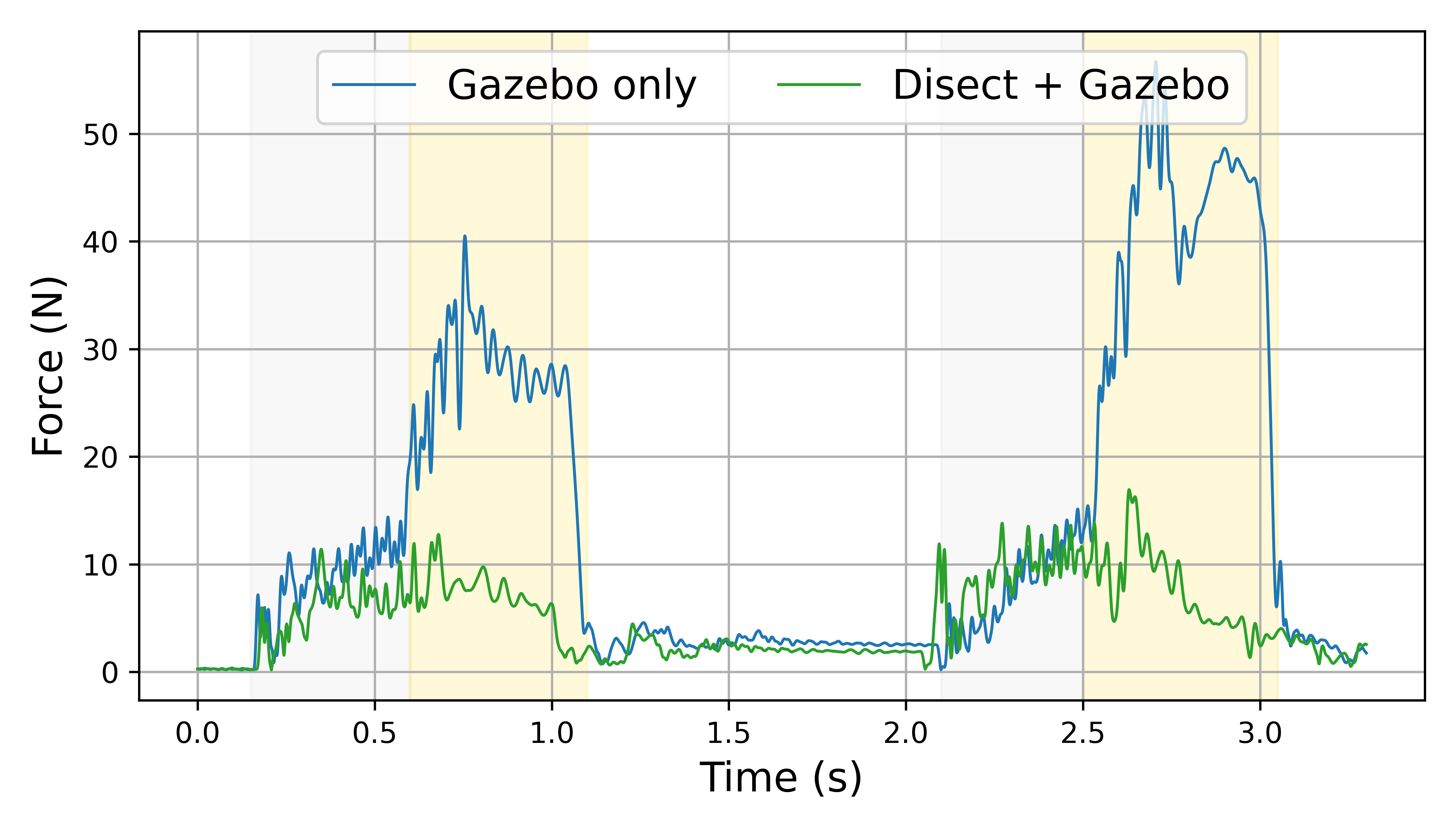}
    \caption{Evaluation on the real robot: Normal contact force at the knife while slicing the cucumber twice. The gray region corresponds to slicing the vegetable, while the yellow region shows the contact with the cutting board.}
    \label{fig:cucumber-wrench}
\end{figure}

\Cref{fig:exp-boxplot} and \ref{fig:cucumber-wrench} illustrate the obtained results. In \Cref{fig:exp-boxplot}, we compare the contact force observed during the slicing actions for each food item between our method and the baseline. These results include the contact force exerted not only on the food item but also on the cutting board. Notably, our method consistently outperforms the baseline by applying significantly lower force during the execution of slicing actions across all tasks. Remarkably, even in the case of the carrot, which was not part of the training dataset, our proposed method demonstrated superior performance when compared to the baseline.

\Cref{fig:cucumber-wrench} centers on the experiments related to cucumber slicing. In this visualization, the grey region corresponds to the slicing of the vegetable, while the yellow region represents the contact force applied to the cutting board.  These findings suggest that the policy acquired through our proposed method achieves superior performance by exhibiting a more adept response to the abrupt stiffness transition between the vegetable and the cutting board.
Similar results are observed across all tasks. These results indicate that the policy learned with our proposed method performed better by more skillfully reacting to the sudden change of stiffness, between the vegetable and the cutting board.
% noise applied to softness and stiffness
\Cref{fig:cucumber-wrench} centers on the experiments related to cucumber slicing. In this visualization, the grey region corresponds to the slicing of the vegetable, while the yellow region represents the contact force applied to the cutting board. The results show clearly that our proposed method performs better than the baseline by more skillfully reacting to the change of stiffness between the food item and the cutting board. These results are consistent across all trials and tasks.

% Optionally \section{Discussion}
\section{Conclusion} \label{sec:conclusion}
% Briefly explain what is our proposed method/system
In this study, we introduced SliceIt! a learning-based robotic system for handling food-cutting tasks. 
Our system combines two key components: a dual simulation environment, and a control policy based on Reinforcement Learning. The aim is to enable a collaborative robot (cobot) or industrial robot arms to perform these tasks safely and accurately by adapting to varying conditions using compliance control.
% Discuss the potential of the proposed system based on the results from the experiments
One of the advantages of using our proposed method is the reduction of food waste while learning the control policies, as only a few real-world samples are required. The experimental results support our hypothesis that using a highly realistic simulation environment is beneficial to learning safer control policies.
% Discuss the current limitations and how they could be addressed in future research work 
 
One limitation of our approach is the increased computation time when using a highly realistic cutting simulator compared to a simplified one. In our experiments, training the RL policy using our method took approximately 40 hours in total, whereas using Gazebo alone required only about 4 hours. This discrepancy arises from the more demanding computational requirements for each simulation time step in DiSECt. However, the additional computation time proves to be worthwhile, given the significant improvement in real robot performance achieved. 
A promising area for future research involves finding ways to reduce the extensive computational time.

\bibliographystyle{IEEEtran}
\bibliography{ref.bib}

\end{document}